%% file: objectness_knowledge_transfer.tex
\documentclass[10pt,twocolumn,letterpaper]{article}

\usepackage{cvpr}
\usepackage{times}
\usepackage{epsfig}
\usepackage{graphicx}
\usepackage{amsmath}
\usepackage{amssymb}
\usepackage{diagbox}
\usepackage{multirow}
\usepackage{color}
\usepackage{array}
\usepackage[normalem]{ulem}
\usepackage{verbatim}

\usepackage[pagebackref=true,breaklinks=true,letterpaper=true,colorlinks,bookmarks=false]{hyperref}

\iftrue  
\newcommand{\stefan}[1]{\textcolor{blue}{[SP: #1]}}
\newcommand{\jasper}[1]{\textcolor{magenta}{#1}}
\newcommand{\vitto}[1]{\textcolor{red}{[VF: #1]}}

\else
\newcommand{\stefan}[1]{\noindent}
\newcommand{\jasper}[1]{\noindent}
\newcommand{\vitto}[1]{\noindent}
\fi
\newcommand{\mypartop}[1]{\vspace{0mm}\paragraph{#1}}
\newcommand{\mypar}[1]{\vspace{-3mm}\paragraph{#1}}

\newcommand{\argmax}{\mathop{\mathrm{argmax}}}

\cvprfinalcopy
\def\cvprPaperID{3739} 

\ifcvprfinal\fi

\begin{document}

\title{Revisiting knowledge transfer for training object class detectors}

{\small
\author{Jasper R. R. Uijlings\\
{\tt\small jrru@google.com}
\and S. Popov\\
{\tt\small spopov@google.com}\\
\small{Google AI Perception}\\
\and V. Ferrari\\
{\tt\small vittoferrari@google.com}
}
}

\maketitle
\thispagestyle{empty}


\input{abstract}

\input{introduction}

\input{method}

\input{results}

\input{results_comparison}

\input{results_comparison_yolo}

\input{result_examples}

\input{results_accross_datasets}
\input{conclusion}

\clearpage
{\small
\bibliographystyle{ieee}
\bibliography{shortstrings,loco}
}

\end{document}

%% file: abstract.tex
\begin{abstract}
We propose to revisit knowledge transfer for training object detectors on target classes
from weakly supervised training images,
helped by a set of source classes with bounding-box annotations.
We present a unified knowledge transfer framework based on
training a single neural network multi-class object detector over all source classes, organized in a
semantic hierarchy. This generates proposals with scores at multiple levels in the hierarchy, which
we use to explore knowledge transfer over a broad range of generality, ranging from class-specific (bicycle to motorbike) to class-generic (objectness to any class). 
Experiments on the 200 object classes in the ILSVRC 2013 detection dataset show that our technique 
(1) leads to much better performance on the target classes (70.3\% CorLoc, 36.9\% mAP) than a weakly supervised baseline which uses manually engineered objectness~\cite{dollar14eccv} (50.5\% CorLoc, 25.4\% mAP).
(2) delivers target object detectors reaching 80\% of the mAP of their fully supervised counterparts.
(3) outperforms the best reported transfer learning
results on this dataset (+41\% CorLoc and +3\% mAP over~\cite{hoffman16jmlr,tang16cvpr},
+16.2\% mAP over~\cite{redmon17cvpr}).
Moreover, we also carry out several across-dataset knowledge transfer
experiments~\cite{lin14eccv,openimages_v2,russakovsky15ijcv} and find that
(4) our technique outperforms the weakly supervised baseline in all dataset pairs by $1.5\times-1.9\times$, establishing its general applicability.

\end{abstract}

%% file: introduction.tex
\section{Introduction}

Recent advances such as \cite{he15cvpr,liu16eccv,ren15nips,zeng16eccv} have resulted in
reliable object class detectors, which predict both the class label and the location of objects in an image.
Typically, detectors are trained under full supervision, which requires manually drawing object bounding-boxes in a large number of training images. This is tedious and very time-consuming.
Therefore, several research efforts have been devoted to training object detectors under weak supervision, i.e. using only image-level labels \cite{bilen14bmvc,bilen16cvpr,cinbis16pami,deselaers10eccv,kantorov16eccv,nguyen09iccv,russakovsky12eccv,siva11iccv,song14icml,song14nips,wang15tip}.
While this is substantially cheaper, the resulting detectors typically perform considerably worse than their fully supervised counterparts. 

\begin{figure*}[thpb]
  \vspace{-.7cm}
  \centering
  \includegraphics[width=\linewidth]{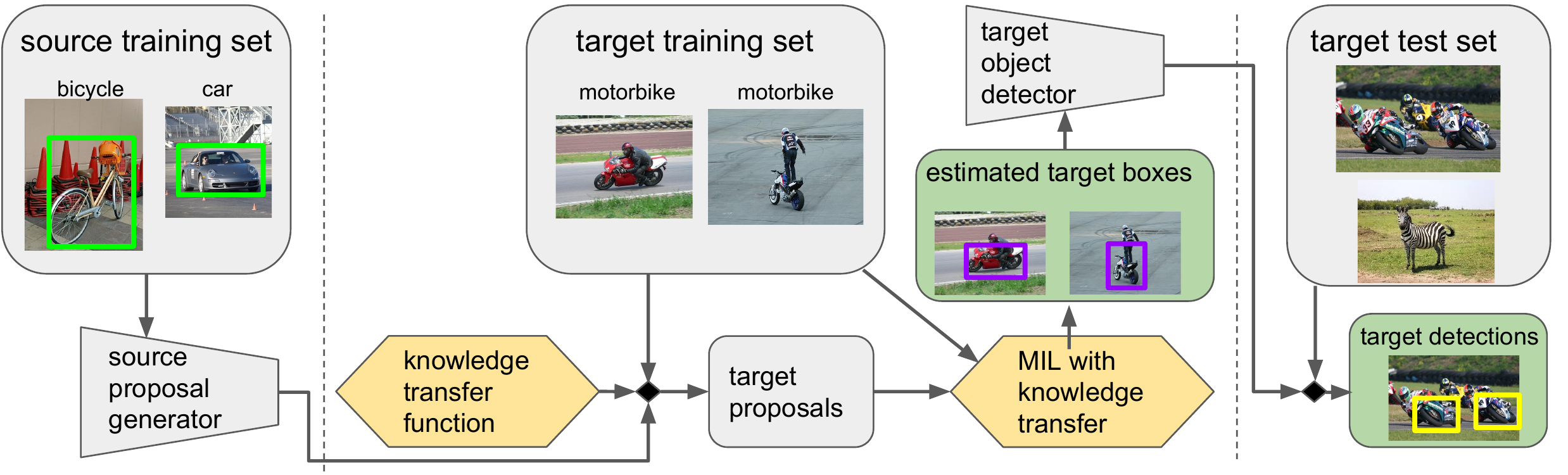}
  \caption{Illustration of our settings and framework. The source training set is
    annotated with bounding-boxes. We use this set to train a proposal generator, and then apply it to the target training set, where it produces proposals with scores at different levels of generality, ranging from class-specific to class-generic. We use these to perform MIL with knowledge transfer (Sec.~\ref{cptTransferLearning}) on the target training set, using only image-level labels (no bounding-boxes). MIL produces boxes for the target classes, which we use to train an object detector. Finally, we apply the object detector to
    the target test set (using no labels at all).
    The performance of our framework is measured
    both on the target training set (Sec.~\ref{ssMIL-results}) and the target test set (Sec.~\ref{ssTest-results}).}
    \label{figPipeline}
    \vspace{-.3cm}
\end{figure*}

In recent years a few large datasets such as ImageNet~\cite{russakovsky15ijcv} and COCO~\cite{lin14eccv}
have appeared, which provide many bounding-box annotations for a wide variety of classes.
Since many classes share visual characteristics, we can leverage these annotations when learning a new class.
In this paper we propose a technique for training object detectors in a knowledge transfer
setting~\cite{guillaumin12cvpr,hoffman16jmlr,redmon17cvpr,rochan2015cvpr,shi12bmvc,tang16cvpr}:
we want to train object detectors for a set of {\em target classes} with only image-level labels, helped by a set of {\em source classes} with bounding-box annotations.
We build on Multiple Instance Learning (MIL), a popular framework for weakly supervised object localization~\cite{nguyen09iccv,deselaers10eccv,bilen14bmvc,cinbis16pami,dietterich97ai,siva11iccv,song14nips,russakovsky12eccv}, and extend it to incorporate knowledge from the source classes.
In standard MIL, images are decomposed into object proposals~\cite{alexe10cvpr,uijlings13ijcv,dollar14eccv} and the process iteratively alternates between re-localizing objects given the current detector, and re-training the detector given the current object locations. During re-localization, typically the highest-scoring proposal for an object class is selected in each image containing it.

Several weakly supervised object localization techniques~\cite{deselaers10eccv,cinbis16pami,prest12cvpr,shapovalova12eccv,siva11iccv,shi12bmvc,tang14cvpr,wang14eccv-cosegmentation,bilen16cvpr} 
incorporate a {\em class-generic} objectness measure~\cite{alexe10cvpr,dollar14eccv} during the re-localization stage, to steer the selection towards objects and away from backgrounds.
These works use a manually engineered objectness measure and report an improvement of around 5\% correct localizations. 
As~\cite{deselaers12ijcv} argued, using objectness can be seen as a (weak) form of knowledge transfer, from a generic object appearance prior to the particular target class at hand.
On the opposite end of the spectrum, several works perform {\em class-specific} transfer~\cite{guillaumin12cvpr,rochan2015cvpr,hoffman16jmlr,tang16cvpr},
For each target class, they determine a few most related source classes to transfer from.
Some works~\cite{guillaumin12cvpr,rochan2015cvpr} use the appearance models of the source classes to guide the localization of the target class by scoring proposals with it, similar to the way objectness is used above.
Other works~\cite{hoffman16jmlr,tang16cvpr} instead perform transfer directly on the parameters of a neural network.
They first train a neural network for whole-image classification on all source and target classes, then fine-tune the source classifiers into object detectors, and finally transfer the parameter transformation between whole-image classifiers and object detectors from the source to related target classes, effectively turning them into detectors too.

Finally, YOLOv2~\cite{redmon17cvpr} jointly trains the source and target class detectors by
combining a standard fully supervised loss with a weakly supervised loss (i.e. the highest scored
box is considered to be the target class). During training they use hierarchical
classification~\cite{cesa06icml,koller97icml,silla11dmkd}, which implicitly achieves knowledge
transfer somewhere in-between class-generic and class-specific.

In this paper we propose a unified knowledge transfer framework for weakly supervised object
  localization which enables us to explore the complete range of semantic specificity (Fig.~\ref{figPipeline}).
We train a single neural network multi-class object detector~\cite{liu16eccv} over all source classes, organized in a semantic hierarchy~\cite{russakovsky15ijcv}.
This naturally provides high-quality proposals and proposal scoring functions at multiple levels in the hierarchy, 
which we use during MIL on the target classes.
The top-level scoring function for `entity' conceptually corresponds to the objectness measure~\cite{alexe10cvpr,dollar14eccv}, but it is stronger, as provided by a neural network properly trained on thousands of images.
Compared to previous works using
objectness~\cite{deselaers10eccv,cinbis16pami,prest12cvpr,shapovalova12eccv,siva11iccv,shi12bmvc,tang14cvpr,wang14eccv-cosegmentation,bilen16cvpr},
this leads to much larger performance improvements on the target classes.
Compared to other transfer
works~\cite{guillaumin12cvpr,redmon17cvpr,rochan2015cvpr,hoffman16jmlr,tang16cvpr},
our framework enables to explore a broad range of generality of transfer:
from the class-generic `entity' class, from intermediate-level categories such as `animal' and `vehicle', and from specific classes such as `tiger' and `car'.
We achieve this in a simple unified framework, using a single SSD model as source
knowledge, where we select the proposal scoring function to be used depending on the target class and the desired level of generality of transfer.

Through experiments on the 200 object classes in the ILSVRC 2013 detection dataset, we demonstrate
that:
(1) knowledge transfer at any level of generality substantially improve results, with class-generic transfer working best. This is excellent
news for practitioners, as they can get strong improvements with a relatively simple modification to
standard MIL pipelines.
(2) our class-generic knowledge transfer leads to large improvements over a weakly supervised object localization baseline using manually engineered objectness~\cite{dollar14eccv}:
70.3\% CorLoc vs 50.5\% CorLoc on the target training set, 36.9\% mAP vs 25.4\% mAP on the target test set.
(3) our method delivers detectors for the target classes which reach 80\% of the mAP of their fully supervised counterparts, trained from manually drawn bounding-boxes.
(4) we outperform the best reported transfer learning results on this dataset:
+41\% CorLoc and +3\% mAP over~\cite{hoffman16jmlr,tang16cvpr},
+16.2\% mAP over~\cite{redmon17cvpr}.
Moreover, we also carry out several across-dataset~\cite{lin14eccv,openimages_v2,russakovsky15ijcv}
knowledge transfer experiments and find that (5) our technique outperforms the weakly supervised
baseline in all dataset pairs by a factor $1.5\times-1.9\times$, establishing its general
applicability.

%% file: method.tex
\section{Method}

We now present our technique for training object detectors in a knowledge transfer setting (Fig.~\ref{figPipeline}).
In this setting we have a training set $\mathcal{T}$ of {\em target classes} with only image-level labels,
and a training set $\mathcal{S}$ of {\em source classes} with bounding-box annotations.
The goal is to train object detectors for the target classes, helped by knowledge from the source classes.

We start in Section \ref{cptMIL} by introducing a reference Multiple Instance Learning (MIL) framework, typically used in weakly supervised object localization (WSOL), i.e. when given only the target set $\mathcal{T}$.
In Section~\ref{cptTransferLearning} we then explain how we incorporate knowledge from the source classes $\mathcal{S}$ into this framework.
Finally, Section~\ref{cptKnowledgeTrafoFcts} discusses the broad range of levels of transfer that we explore.

\subsection{Reference Multiple Instance Learning (MIL)}
\label{cptMIL}

\mypartop{General scheme.}
For simplicity, we explain here MIL for one target class $t \in \mathcal{T}$. The process can be repeated for each target class in turn.
The input is a training set $\mathcal{I}$ with positive images, which contain the target class, and negative images, which do not. Each image is represented as a bag of object proposals $\mathcal{B}$ extracted by a generator such as~\cite{alexe10cvpr,dollar14eccv,uijlings13ijcv}.
A negative image contains only negative proposals, while a positive image contains at least one positive proposal, mixed in with a majority of negative ones. The goals are to find the true positive proposals and to learn an appearance model $A_t$ for class $t$ (the object detector). This is solved in an iterative fashion, by alternating between two steps until convergence:
\begin{enumerate}
  \item {\bf Re-localization:} in each positive image $I$, select the proposal $b^*$ with the highest score given by the current appearance model $A_t$:
\begin{equation}
    \label{eqMIL}
    b^* \equiv \argmax_{b \in \mathcal{B}} A_t(b, I)
\vspace{-0.2cm}
\end{equation}
  \item {\bf Re-training:} re-train $A_t$ using the current selection of proposals from the positive images, and all proposals from negative images.
\end{enumerate}
\vspace{-0.1cm}
\mypar{Features and appearance model.}
Typical MIL implementations use as appearance model a linear SVM trained on CNN-features extracted from the object proposals~\cite{girshick14cvpr,cinbis16pami,bilen14bmvc,bilen15cvpr,song14icml,wang15tip}.

\mypar{Initialization}
In the first iteration many works train the appearance model by using complete images minus a small boundary as positive training examples~\cite{cinbis14cvpr,cinbis16pami,pandey11iccv,russakovsky12eccv,nguyen09iccv,kim2009nips}.

\mypar{Multi-folding.}
In a high dimensional feature space the SVM separates positive and negative training examples well, placing most positive samples far from the decision hyperplane.
Hence, during re-localization the detector is likely to score the highest on the object proposals which were used as positive training samples in the previous iteration.
This leads MIL to get stuck on some incorrect selection in early iterations.
To prevent this,~\cite{cinbis16pami} proposed multi-folding: the training set is split into 10 subsets, and then the re-localization on each subset is done using detectors
trained on the union of all other subsets.

\mypar{Objectness.}
Objectness was proposed in~\cite{alexe10cvpr} to measure how likely it is that a proposal tightly encloses an object of any class (e.g. bird, car, sheep), as opposed to background (e.g. sky, water, grass).
Since~\cite{deselaers10eccv}, many WSOL techniques~\cite{bilen16cvpr,cinbis16pami,prest12cvpr,shapovalova12eccv,siva11iccv,shi12bmvc,tang14cvpr,wang14eccv-cosegmentation} have used an objectness measure~\cite{alexe10cvpr,dollar14eccv} to steer the re-localization process towards objects and away from background. 
Following standard practice, incorporating objectness into Eq~\eqref{eqMIL} leads to:
\begin{equation}
    \label{eqMILObjectness}
    b^* \equiv \argmax_{b \in \mathcal{B}} \lambda \cdot A_t(b, I) + (1 - \lambda) \cdot O(b,I)
\end{equation}
where $\lambda$ is a weight controlling the trade-off between the class-generic objectness score $O$ and the appearance model $A_t$ of the target class $t$ being learned during MIL.
Using the objectness score in this manner, previous works typically report an improvement around 5\%
in correct localizations of the target objects~\cite{bilen16cvpr,deselaers10eccv,cinbis16pami,prest12cvpr,shapovalova12eccv,siva11iccv,shi12bmvc,tang14cvpr,wang14eccv-cosegmentation}.

\subsection{MIL with knowledge transfer}
\label{cptTransferLearning}

\begin{figure}[t]
  \centering
  \vspace{-.5cm}
  \includegraphics[width=\linewidth]{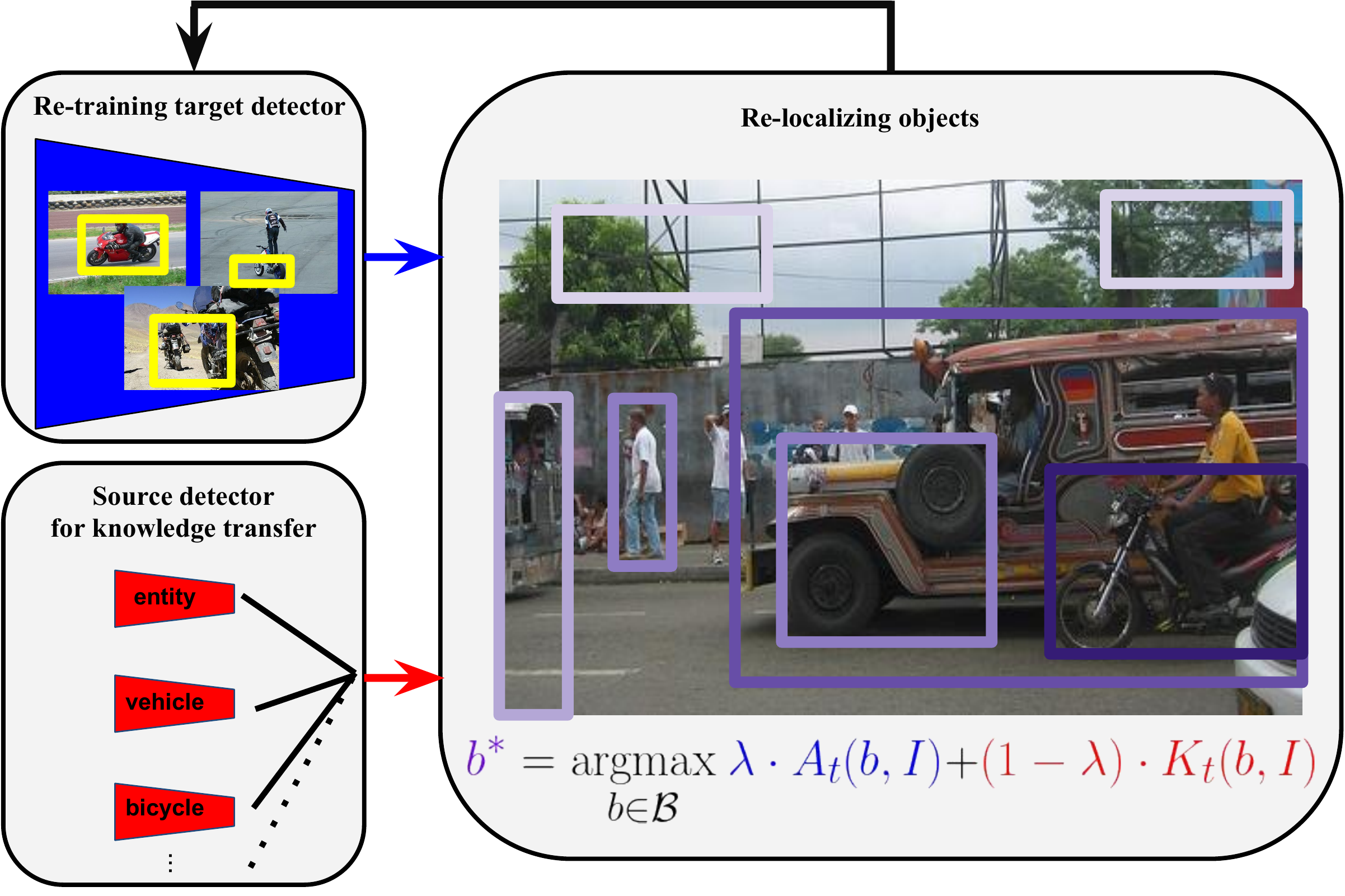}
  \caption{Illustration of MIL + knowledge transfer for the target class
    `motorbike'. Standard MIL consists of a re-training stage and a
    re-localization stage (Sec.~\ref{cptMIL}). We add knowledge transfer to this scheme by training
    SSD~\cite{liu16eccv} on the hierarchy $\mathcal{H}$ defined by the source set $\mathcal{S}$. We
    use its proposals and a knowledge transfer function (Sec.~\ref{cptKnowledgeTrafoFcts}) in the
  re-localization stage.}
  \label{figFramework}
  \vspace{-.3cm}
\end{figure}

\mypartop{Overview.}
In the strandard WSOL setting, MIL is applied only on the training set $\mathcal{T}$ of target classes with image-level labels.
In our setting we also have a training set of source classes $\mathcal{S}$ with bounding-box annotations.
Therefore we incorporate knowledge from the source classes into MIL and help learning detectors for the target classes (Fig.~\ref{figPipeline}).
We train a multi-class object detector over all source classes $\mathcal{S}$ organized in a
semantic hierarchy, and then apply it to $\mathcal{T}$ as a proposal generator. This naturally
provides high-quality proposals, as well as proposal scoring functions at multiple levels in the
hierarchy. We use these scoring functions during the re-localization stage of MIL on $\mathcal{T}$
(Fig.~\ref{figFramework}), which greatly helps localizing target objects correctly (Sec.~\ref{cptResults}).

Our scheme improves over previous usage of manually engineered object proposals and their associated objectness score~\cite{alexe10cvpr,dollar14eccv} in WSOL in two ways:
(1) by using {\em trained} proposals and objectness scoring function trained on source classes;
(2) generalizing the common use of a single class-agnostic objectness score to a {\em family of proposal scoring functions} at multiple levels of semantic specificity.
This enables exploring using scoring functions tailored to particular target classes, and at
various degrees of relatedness between source and target classes (Sec.~\ref{cptKnowledgeTrafoFcts}).
Below we explain the elements of our approach in more detail.

\mypar{Training a proposal generator on the source set.}
We use the Single Shot Detection (SSD) network~\cite{liu16eccv}.
SSD starts from a dense grid of `anchor boxes' covering the image, and then adjusts
their coordinates to match objects using regression. This in turn enables substituting
Region-of-Interest pooling~\cite{he14eccv,girshick15iccv,ren15nips} with convolutions, yielding
considerable speed-ups at a small loss of performance~\cite{huang17cvpr}.  The SSD implementation we
use has Inception-V3~\cite{szegedy16cvpr} as base network and outputs 1296 boxes per image.

\begin{figure}[t]
  \centering
  \includegraphics[width=\linewidth]{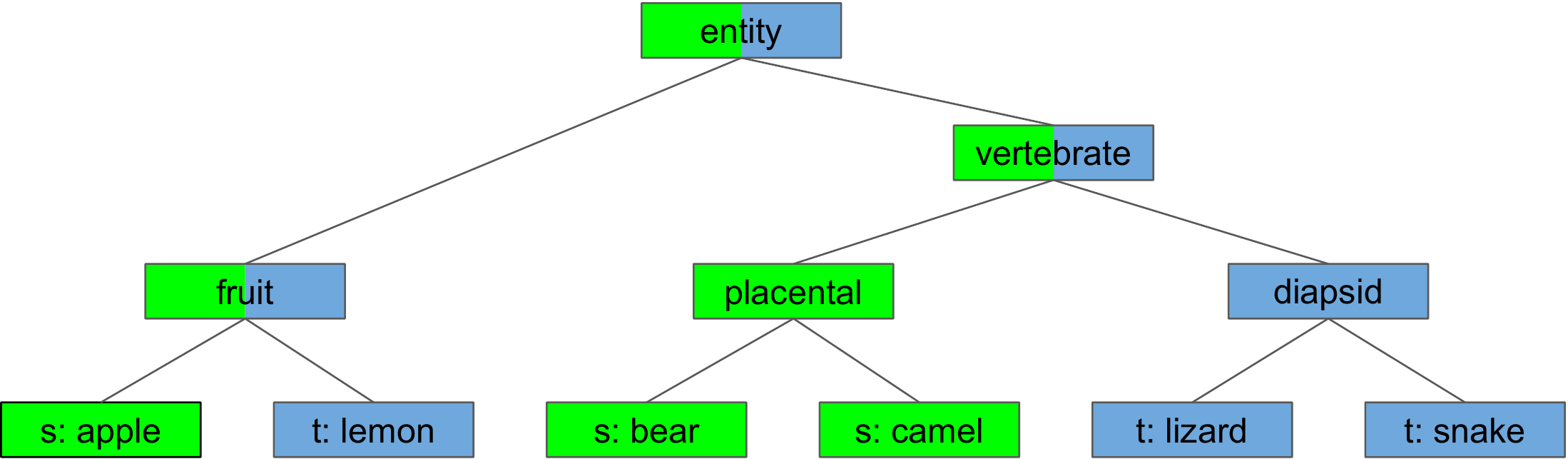}
  \caption{Illustration of part of the ImageNet hierarchy, with our source and target classes inside it.
    Source classes in $\mathcal{S}$ are the leaf nodes in green. Target classes in $\mathcal{T}$ are leaf nodes in blue.
    For other nodes the color shows whether it has only source classes as leaves under it, only
  target classes, or a mixture of both.}
  \label{figHierarchy}
\end{figure}

We train SSD on the source set $\mathcal{S}$. For each anchor box, SSD regresses to a single output
box, along with one confidence score for each source class.  Therefore, the proposal set
$\mathcal{B}$ generated for an image is class-generic.
Before training SSD, we first position the source classes $\mathcal{S}$ into the ImageNet semantic
hierarchy $\mathcal{H}$~\cite{russakovsky15ijcv} and expand the label space to include all ancestor
classes up to the top-level class `entity' (Fig.~\ref{figHierarchy}).  After this expansion, each
object bounding-box has multiple class labels, including its original label from $\mathcal{S}$ (e.g.
`bear') and all its ancestors up to `entity' (e.g. `placental', `vertebrate', `entity').
Hence, we train SSD in a multi-label setting, and we use a sigmoid cross entropy loss for each class
separately, instead of the common log softmax loss across classes (which is suited for standard
1-of-$K$ classification, e.g.~\cite{girshick15iccv,liu16eccv,ren15nips}).
Note how this entails that ancestor classes use as training samples the set union of all samples over their descendants in $\mathcal{S}$.

We stress that we use SSD instead of Fast- or Faster-RCNN detectors~\cite{girshick15iccv,ren15nips}
because these detectors perform class-specific bounding-box regression. That leads to different sets
of boxes for each source class, which complicates experiments in our knowledge transfer setting. SSD
instead delivers a single set of class-generic boxes, and attaches to each box multiple scores (one
per source class).

\mypar{Knowledge transfer during re-localization.}
After training SSD on $\mathcal{S}$, we apply it to each image $I$ in the target set $\mathcal{T}$. It produces a set of proposals $\mathcal{B}$
and assigns to each proposal $b\in \mathcal{B}$ scores $F_s(b, I)$ at all levels of the hierarchy.
More precisely, it assigns a score for each class $s\in \mathcal{H}$, including scores for the original leaf 
classes $\mathcal{S}$, the intermediate-level classes, and the top-level `entity' class.
This top-level score corresponds conceptually to traditional
objectness~\cite{alexe10cvpr,dollar14eccv}, but is now properly trained.

We use this family of scoring functions $F_s$ to compose one particular knowledge transfer function
$K_t(b, I)$ tailored to each target class $t \in \mathcal{T}$. We discuss in Sec.~\ref{cptKnowledgeTrafoFcts}
four strategies for composing $K_t$.
As illustrated in Fig.~\ref{figFramework}, we use $K_t$ inside the re-localization stage of MIL
by generalizing Eq~\eqref{eqMILObjectness} to become:
\begin{equation}
    \label{eqMILKnowledge}
    b^* \equiv \argmax_{b \in \mathcal{B}} \lambda \cdot A_t(b,I) + (1 - \lambda) \cdot K_t(b,I)
\end{equation}
Note how the special case of $K_t(b,I) = O$ (using a standard objectness score~\cite{dollar14eccv,alexe10cvpr}) and $\mathcal{B}$ coming from a standard object proposal generator corresponds to WSOL with the reference MIL algorithm (sec.~\ref{cptMIL}).

\subsection{Knowledge transfer functions $K_t$}
\label{cptKnowledgeTrafoFcts}

In this paper we explore knowledge transfer at different levels of semantic generality.
For a given target class $t$, we use the proposal scoring functions $F_s$ to compose a knowledge transfer function $K_t$ at a desired levels of generality, out of four possible options: class-generic, closest source classes, closest common ancestor, and closest common ancestor with at least $n$ sources.

\mypartop{Class-generic objectness.}
The most generic way of transferring knowledge is to use the scoring function
$F_{\textrm{entity}}(b,I)$ from the top-level class `entity' in the hierarchy.
The idea is that such a generic measure generalizes beyond the source classes it was trained on, and it helps steer the re-localization process towards objects and away from background in the target dataset. This corresponds to the traditional use of objectness in
WSOL~\cite{deselaers10eccv,cinbis16pami,prest12cvpr,shapovalova12eccv,siva11iccv,shi12bmvc,tang14cvpr,wang14eccv-cosegmentation,bilen16cvpr},
but done with a stronger objectness measure trained in a neural network:
\begin{equation}
  K_t(b,I) \equiv F_{\textrm{entity}}(b,I)
    \label{eqKnowledgeGeneric}
\end{equation}
In our main experiments, this scoring function is trained on the set union over the training samples of all $100$ source classes in the dataset we use (Sec.~\ref{cptResults}).

\mypar{Closest source classes.}
On the other end of the spectrum, we can transfer knowledge from the most similar source classes to
the target $t$, the most common scenario in a knowledge transfer setting~\cite{guillaumin12cvpr,hoffman16jmlr,rochan2015cvpr,shi12bmvc,tang16cvpr}. To find these source classes, we consider the position of $t$ in the semantic
hierarchy $\mathcal{H}$.  We find the closest ancestor $a_1$ of $t$ whose descendants include at least one
source class from $\mathcal{S}$, and then take all leaf-most source classes among its descendants. Often these
closest sources are the siblings of $t$ (e.g. lemon for apple in Fig.~\ref{figHierarchy}), but if
$t$ has no siblings in $\mathcal{S}$ the procedure backtracks to a higher-level ancestor and takes
its descendants instead (e.g. bear and camel for lizard in Fig.~\ref{figHierarchy}).
In practice, a target class has a median number of $3$ closest source classes in our main experiments (Sec.~\ref{cptResults}).

We combine the scoring functions of the closest source classes $\mathcal{N}_t$ into the knowledge transfer function as:
\begin{equation}
  K_t(b,I) \equiv \frac{1}{|\mathcal{N}_t|} \sum_{s \in \mathcal{N}_t} F_s(b,I)
    \label{eqKnowledgeSpecific}
\end{equation}

\mypar{Closest common ancestor.}
A different way to use the training data from the closest source classes is to directly use the scoring function $F_{a_1}$ of the closest ancestor $a_1$ of $t$ who has descendants in $\mathcal{S}$:
\begin{equation}
    K_t(b,I) \equiv F_{a_1}(b,I)
    \label{eqKnowledgeAncestor}
\end{equation}
The scoring function $F_{a_1}$ is trained from the set union of the training data over all closest
source classes $\mathcal{N}_t$ (instead of averaging the scoring function outputs in Eq.~\eqref{eqKnowledgeSpecific}).

\mypar{Closest common ancestor with at least $n$ sources.}
The extreme cases above present a trade-off. The `entity' class is trained from a lot of samples,
but is very generic. In contrast, the closest source classes are more specific to the target, but
have less training data to form strong detectors.

Here we propose an intermediate approach: we control the degree of generality of transfer by setting a minimum to the number of source classes an ancestor should have.
We define $a_n$ as the closest ancestor of $t$ who has at least $n$ source classes as descendants. This generalizes Eq.~\eqref{eqKnowledgeAncestor} to:
\begin{equation}
    K_t(b,I) \equiv F_{a_n}(b,I)
    \label{eqKnowledgeAncestorGeneralized}
\end{equation}
Note that setting $n=|\mathcal{S}|$ leads to to selecting the \emph{entity} class as ancestor, matching Eq.~\eqref{eqKnowledgeGeneric}.
In our experiments in Sec.~\ref{cptResults} we set $n=5$, resulting in a median of $10$ source classes under the ancestor selected for each target class.

%% file: results.tex
\section{Results on ILSVRC 2013}
\label{cptResults}

\begin{table*}[t]
  \vspace{-.7cm}
  \footnotesize
  \setlength\tabcolsep{4.7pt}
  \begin{tabular}{|l|c|c|c|c|c||c|c||c|}
    \hline
    & \multicolumn{5}{c||}{Knowledge transfer results} & \multicolumn{2}{c||}{EdgeBoxes baseline} & Full
    supervision \\
    \hline
    Knowledge transfer function $K_t$ & none &  closest sources & closest ancestor & ancestor $a_5$
    & class-generic & none & objectness & -\\
    \hline
    \hline
    CorLoc IoU $> 0.5$ & 41.4 & 68.5 & 68.4 & 68.6 & {\bf 70.3} & 39.4 & 50.5 & - \\
    CorLoc IoU $> 0.7$ & 20.0 & 56.4 & 56.4 & 56.6 & {\bf 58.8} & 18.4 & 28.5 & - \\
    \hline
    mAP IoU $>0.5$ & 23.2 & 35.3 & 34.8 & 35.9 & {\bf 36.9} & 20.7 & 25.4 & 46.2 \\
    mAP IoU $>0.7$ & 7.0 & 25.7 & 25.5 & 25.9 & {\bf 27.2} & 6.7 & 11.0 & 31.7 \\
    \hline
  \end{tabular}
  \caption{Results for various knowledge transfer functions and full supervision on the
    target training set (CorLoc) and target test set (mAP). Knowledge transfer results significantly
    outperform the MIL baseline (EdgeBoxes). Class-generic knowledge transfer works best.}
  \label{tabResults}
  \vspace{-.1cm}
\end{table*}

\mypartop{Dataset.}
We use ILSVRC 2013~\cite{russakovsky15ijcv}, following exactly the same settings
as~\cite{hoffman16jmlr,tang16cvpr} to enable direct comparisons.
We split the ILSVRC 2013 validation set into two subsets val1 and val2, and augment val1 with images from the ILSVRC 2013 training set such that each class has 1000 annotated bounding-boxes in total~\cite{girshick14cvpr}.
ILSVRC 2013 has 200 object classes: we use the first 100 as sources $\mathcal{S}$ and second 100
as targets $\mathcal{T}$ (classes are alphabetically sorted).

As our \emph{source training set} we use all images of the augmented val1 set which have
bounding-box annotations for 100 source classes, resulting in 63k images with 81k bounding-boxes.
As \emph{target training set} we use all images of the augmented val1 set which contain the 100 target classes and remove all bounding-box annotations, resulting in 65k images with 93k image-level labels.
In Sec.~\ref{ssMIL-results} we report results for MIL applied to the target training set.
As \emph{target test set} we use all 10k images of val2 and remove all annotations.
In Sec.~\ref{ssTest-results} we train a detector from the output of MIL on the target training set, and evaluate it on the target test set.
Finally, in Sec.~\ref{ssComparison} we compare to three previous
works~\cite{hoffman16jmlr,tang16cvpr,redmon17cvpr}
on knowledge transfer on ILSVRC 2013.

\subsection{Knowledge transfer to the target training set}
\label{ssMIL-results}

We first explore the effects of knowledge transfer for localizing objects in the target training set.

\mypar{Settings for MIL with knowledge transfer.}
We train SSD~\cite{liu16eccv} on the source training set and
apply it to the target training set to produce object proposals and corresponding scores
(Sec.~\ref{cptTransferLearning}). Then we apply MIL on the target training set
(Sec.~\ref{cptTransferLearning}) while varying the knowledge transfer function $K_t$
(Sec.~\ref{cptKnowledgeTrafoFcts}) during re-localization (Eq.~\eqref{eqMILKnowledge}).

Following~\cite{bilen14bmvc,bilen15cvpr,cinbis16pami,song14icml,wang15tip} during MIL we describe
each object proposals with a 4096-dimensional feature vector using the Caffe
implementation~\cite{jia13caffe} of the AlexNet CNN~\cite{krizhevsky12nips}. As customary, we use
weights from~\cite{jia13caffe} resulting from training on ILSVRC
classification~\cite{russakovsky15ijcv} using only image-level labels (no bounding-box annotations).
As appearance model we use a linear SVM on these features.

For each knowledge transfer function, we optimize $\lambda$ in Eq.\eqref{eqMILKnowledge} on the
source training set in a cross-validation manner: we subdivide this set in 80 source classes and 20 target classes, run our knowledge transfer framework, and choose the $\lambda$ which leads to the highest localization performance (CorLoc, see below).

\mypar{Evaluation measure.}
We quantify localization performance with Correct Localization (CorLoc)~\cite{deselaers10eccv} averaged over the target classes $\mathcal{T}$.
CorLoc is the percentage of images of class $t$ where the method correctly localizes one of its
instances. We consider two Intersection-over-Union (IoU)~\cite{everingham10ijcv} thresholds:
we report CorLoc at IoU $> 0.5$ (commonly used in the WSOL literature~\cite{bilen16cvpr,cinbis16pami,deselaers10eccv,tang14cvpr}) and IoU $> 0.7$ (stricter criterion requiring tight localizations).

\mypar{Quantitative results.}
The first two rows of Table~\ref{tabResults} report CorLoc on the target training set.
As a baseline we use no knowledge transfer function at all. This leads to
41.4\% CorLoc at IoU $>0.5$.

All the forms of knowledge transfer we explore yield massive improvements over the baseline:
27-29\% CorLoc increase. Interestingly, simply transferring from the top-level `entity' class works best and yields 70.3\% CorLoc. This shows that the trade-off between semantic generality and number of source
training samples is skewed towards the former. We believe this is excellent news for the
practitioner: our experiments show that a simple modification to standard MIL pipelines can lead to
dramatic improvements in localization performance (i.e. just change the scoring function used during
re-localization to include a strong objectness function trained on the source set).

When measuring CorLoc at the stricter IoU$>0.7$ threshold, the benefits of knowledge transfer are
even more pronounced. The baseline without knowledge transfer only brings 20\% CorLoc, while
class-generic transfer achieves 58.8\% CorLoc, almost $3\times$ higher. This suggests
that knowledge transfer enables localizing objects with tighter bounding-boxes.

\mypar{Reference MIL with manually engineered proposals.}
In the previous experiments we transferred knowledge from the source classes not only via the
knowledge transfer functions, but also by using trained object proposals:
the locations of the proposals produced by SSD on the target training set are influenced by the locations of the objects in the source training set.

To eliminate all forms of knowledge transfer, we perform here experiments using the same MIL
framework as before, but now using untrained, manually engineered EdgeBox
proposals~\cite{dollar14eccv}.  Without using any objectness function during re-localization
(Eq.~\eqref{eqMIL}), this baseline obtains 39.4\% CorLoc at IoU $>0.5$. In contrast, our
SSD proposals without any knowledge transfer function yields 41.4\% CorLoc. This shows that
trained object proposal locations helps only a little.
Furthermore, using also the untrained, class-generic objectness of Edgeboxes during
re-localization (Eq.~\eqref{eqMILObjectness}) results in 50.5\% CorLoc, an improvement of 11\%. In
contrast, our trained class-generic knowledge transfer yields 70.3\% CorLoc, a much higher
improvement of 29\%.
This system (MIL with EdgeBoxes and Objectness) is the reference MIL of Sec.~\ref{cptMIL}, which
represents a standard WSOL method without learned knowledge transfer functions.

The above experiments demonstrate that the major reason for the performance improvement brought by our knowledge transfer scheme is the knowledge transfer functions, not the trained proposals.

\mypar{A closer look at closest sources.}
The previous section showed that class-generic transfer outperforms class-specific transfer in our experiments. As this may seem counter-intuitive, we investigate here whether our closest source strategy could be improved.

Above we used distance in the WordNet hierarchy to find the closest source
classes to a target, which reflects semantic similarity rather than visual similarity. Here we
perform an additional experiment using visual similarity instead.
We extract whole-image features on the source and target training sets using an
AlexNet~\cite{krizhevsky12nips} classification network pre-trained on ILSVRC
classification~\cite{russakovsky15ijcv}. For each class, we average the features of all its training
images. For each target class, the closest source class is the one with the most similar averaged
features, measured in Euclidean distance.

We also compute an upper-bound performance by selecting for each target class the source class that
leads to the highest CorLoc on the target training set. This is the best possible source. Note how
this experiment needs ground-truth bounding-boxes on the target training set to select a source
class, and so it is not a valid strategy in practice. It is only intended to reveal the upper-bound
that any way of selecting a source specific to a target cannot exceed.

The results in Tab.~\ref{tabClosestSources} show that using either semantic similarity or visual similarity yields similar results: 68.5\% and 68.0\% CorLoc respectively.
Using the best source class improves moderately over both automatic ways to select a source class (69.6\% CorLoc). Interestingly, even the best source class does not outperform
class-generic knowledge transfer (70.3\% CorLoc).  This is likely due to the fact that individual source
classes have too little training data to form strong detectors, whereas the class-generic objectness
model benefits from a very large training set (effectively the set union of all sources).

\begin{table}
  \small
  \centering
  \begin{tabular}{|l|c|}
    \hline
    closest source strategy & CorLoc IoU $> 0.5$ \\
    \hline
    WordNet hierarchy 		& 68.5 \\
    Visual similarity 		& 68.0 \\
    Best source upper-bound 	& 69.6 \\
    \hline
    \hline
    class-generic & 70.3 \\
    \hline
  \end{tabular}
  \caption{Knowledge Transfer using different ways to determine the closest source class. Even
  the upper-bound does not outperform class-generic knowledge transfer.}
  \label{tabClosestSources}
\end{table}

\mypar{Correlation between semantic similarity and improvement.}

\begin{figure}
  \vspace{-0.7cm}
\includegraphics[width=\linewidth]{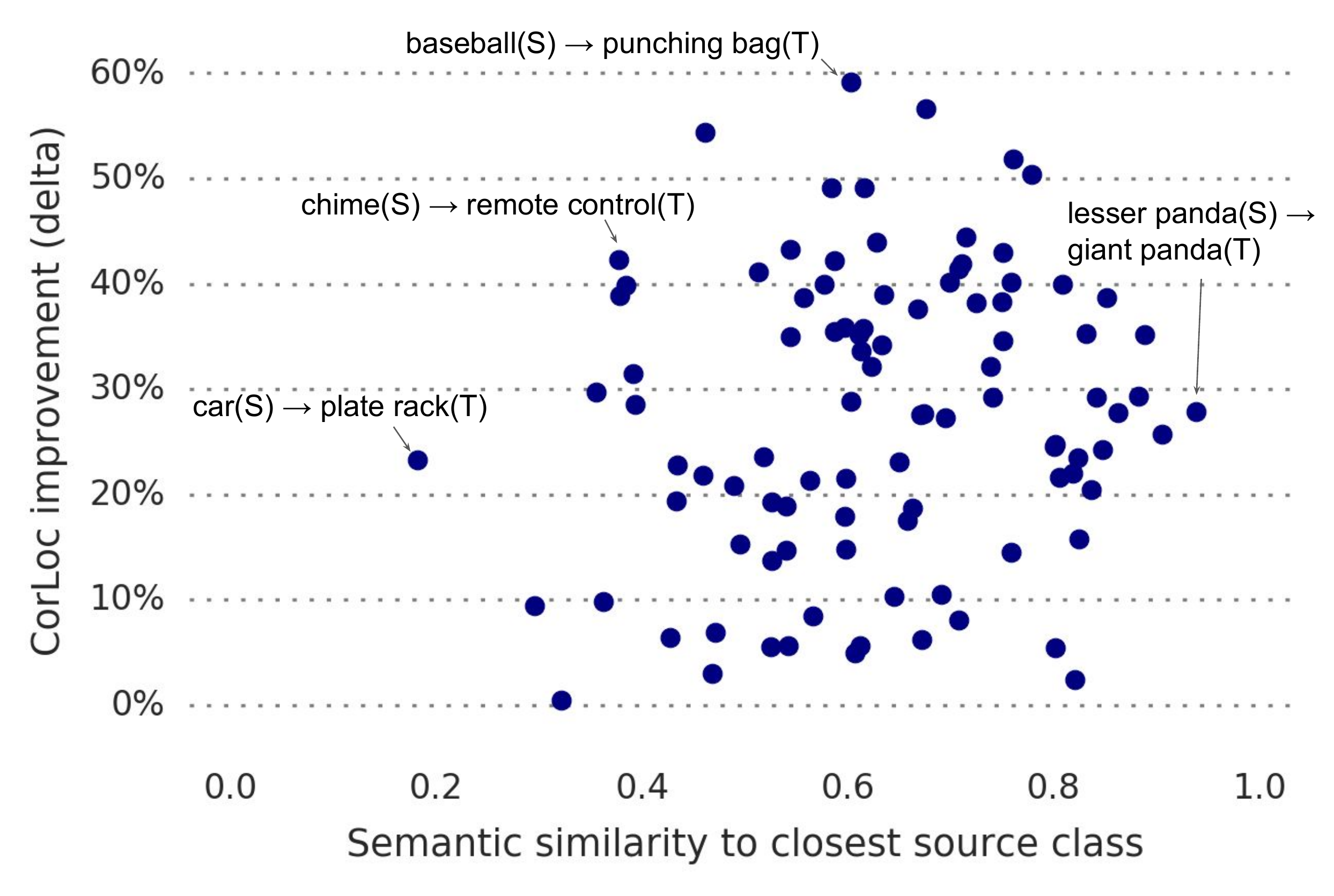}
\caption{
Absolute CorLoc improvement brought by our class-generic knowledge transfer, as a function of semantic similarity~\cite{lin98icml} between a target class and the most similar source class. Each point
  represents one target class. For several points we show which source class (S) it represents, and its most similar class (T).}
\label{figCorLocVsSimilarity}
\end{figure}

We now investigate whether there is a relation between the improvements brought by our class-generic knowledge transfer on a particular target class, and its semantic similarity to the source classes.
We measure semantic similarity by the widely used Lin similarity~\cite{lin98icml} on WordNet (same hierarchy as ImageNet). For each target class in $\mathcal{T}$, the horizontal axis in
Fig.~\ref{figCorLocVsSimilarity} reports the similarity of the most similar source class in
$\mathcal{S}$.  The vertical axis reports the absolute CorLoc improvement on the target training
set, over the no-transfer baseline.

Interestingly, we observe no significant correlation between CorLoc improvement and semantic
similarity. This suggests that this knowledge transfer function, trained on a large set of 100
diverse source classes, is truly class generic.

\subsection{Object detection on the target test set}
\label{ssTest-results}

We now train an object detector from the bounding-boxes produced on the target training set by MIL.
We train a Faster-RCNN detector~\cite{ren15nips} with Inception-ResNet~\cite{szegedy17aaai} as base
network.  We apply it to the target test set and report mean Average
Precision (mAP)~\cite{everingham10ijcv,russakovsky15ijcv}.

As Tab.~\ref{tabResults} shows, mAP on the test set correlates very well with CorLoc on
the training set. At IoU$>0.5$, the best results are brought by our class-generic transfer method
(36.9\% mAP), strongly improving over the no-transfer baseline (23.2\%) and the EdgeBoxes +
objectness baseline (25.4\%).
Results at IoU$>0.7$ reveal an interesting phenomenon: both baselines fail to train an
object detector that localizes objects accurately enough (7.0\% mAP for no-transfer, 11.0\% mAP for
EdgeBoxes + objectness). Instead, our class-generic
knowledge transfer scheme succeeds even at this strict threshold. Its mAP (27.2\%) is around $4
\times$ and $2.5\times$ higher than the baselines.
To put our results in context, we also report mAP when training on the target training set with
ground-truth bounding-boxes, which acts as an upper-bound (`full supervision' in
Tab.~\ref{tabResults}). At IoU $>0.5$ and IoI $>0.7$,
our scheme reaches 80\% and 86\% of this upper-bound, respectively.

These experiments consolidates our findings and shows that our simple class-generic transfer
strategy is effective in improving the performance of object detectors for target classes
for which only image-level labels are available.

%% file: results_comparison.tex
\subsection{Comparison to~\cite{hoffman16jmlr,tang16cvpr,redmon17cvpr}}
\label{ssComparison}

We now compare our technique to two transfer learning works~\cite{hoffman16jmlr,tang16cvpr}
using the exact same dataset with the same source and target training splits as
in~\cite{hoffman16jmlr,tang16cvpr} (see Sec.~\ref{cptResults}).

\begin{table}[t]
  \vspace{-0.7cm}
  \footnotesize
  \centering
  \begin{tabular}{|l|c|c|}
    \hline
    & CorLoc IoU $>$ 0.5 & mAP IoU $>$ 0.5 \\
    \hline
    LSDA~\cite{hoffman16jmlr} & 28.8 & 18.1 \\
    Tang et al.~\cite{tang16cvpr} & - & 20.0 \\
    {\bf our method} & 70.3 & 23.3 \\
    \hline
  \end{tabular}
  \caption{Comparison of our results to~\cite{hoffman16jmlr,tang16cvpr} at IoU $>0.5$. All numbers
  presented in this table use AlexNet~\cite{krizhevsky12nips} as base network. Tang et al.~\cite{tang16cvpr} does not report CorLoc.}
  \label{tabSoA}
\end{table}

In terms of CorLoc on the target training set, LSDA~\cite{hoffman16jmlr} reports 28.8\% at IoU $>0.5$
while our method delivers 70.3\%, more than twice higher (Tab.~\ref{tabSoA}). Note that~\cite{hoffman16jmlr,tang16cvpr} and our MIL method all use the same base network (AlexNet~\cite{krizhevsky12nips}) to produce feature descriptors for proposals. Hence, they are directly comparable.

In terms of performance on the target test set, in order to make a fair comparison
to~\cite{hoffman16jmlr,tang16cvpr}, we train a Fast-RCNN detector model~\cite{girshick15iccv} using
the same base network: AlexNet~\cite{krizhevsky12nips} (as opposed to the results in
Tab.~\ref{tabResults}, which use a stronger detector).
Our method leads to detectors performing at 23.3 mAP on the target test set, improving over the 20.0 by \cite{tang16cvpr} and 18.1 by~\cite{hoffman16jmlr}.
Moreover, our method is also much simpler: just insert a properly trained class-generic objectness scoring function into standard MIL pipelines.

%% file: results_comparison_yolo.tex
We also compare to YOLOv2 following their settings (Section 4 in~\cite{redmon17cvpr}):
COCO train as the source training set, ImageNet-classification as the target training
set, and ILSVRC-detection validation as the target test set. 
In our setup we subsample ImageNet-classification by randomly selecting up to 1K images for each
of the 200 target classes.

In the spirit of knowledge transfer, ~\cite{redmon17cvpr} report results over the 156 target classes that are not present in
COCO. On those classes, our method yields 32.2 mAP, substantially better than the 16.0 mAP of~\cite{redmon17cvpr}.

We note that the object detection model that we use seems approximately comparable to YOLOv2 (Table 3 of~\cite{redmon17cvpr}).
This shows that our improvement is due to better transfer learning.

%% file: result_examples.tex
\begin{figure*}[hbt]
  \vspace{-0.7cm}
  \includegraphics[width=\textwidth]{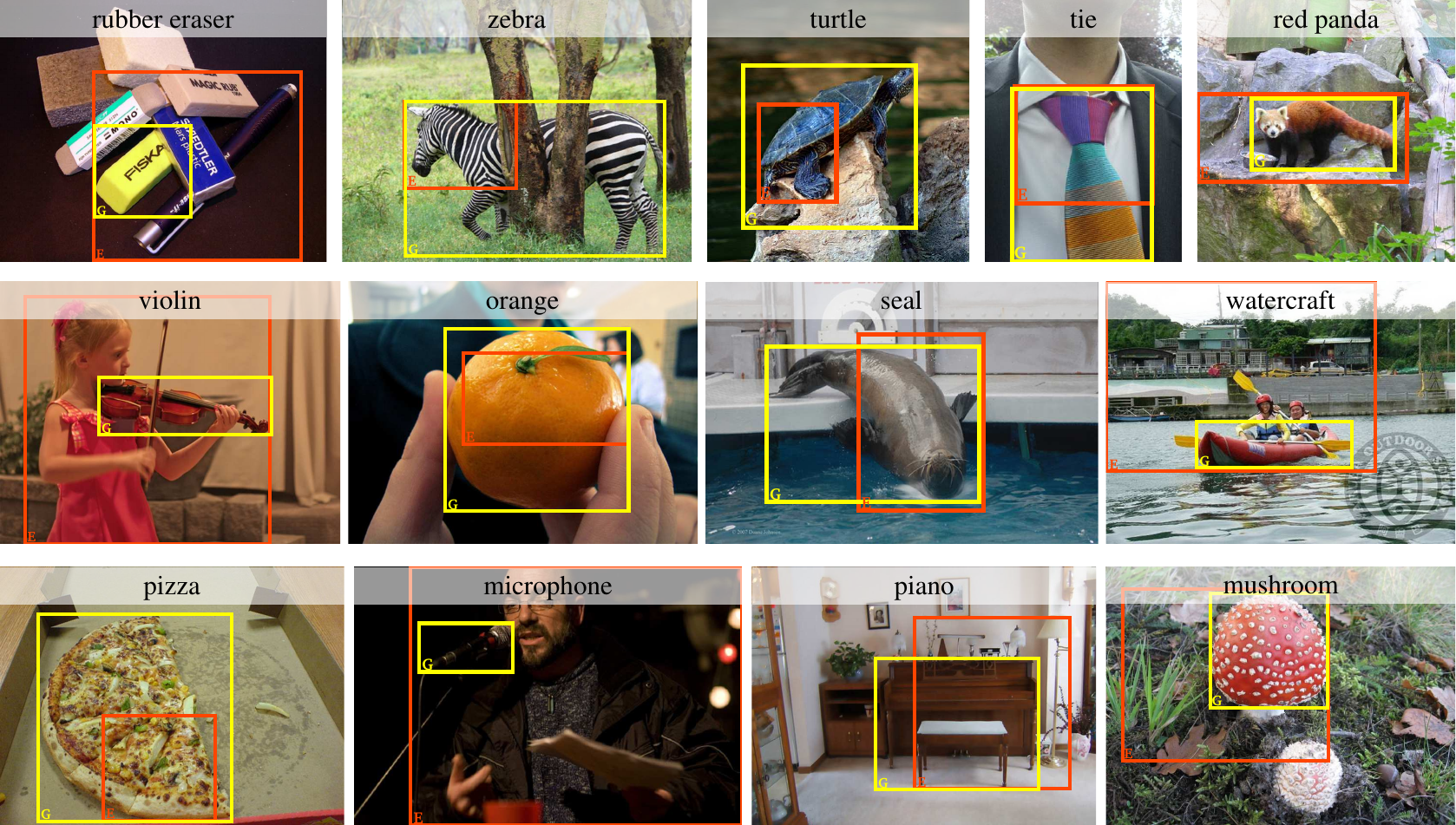}
  \caption{
    Example localizations produced by our class-generic knowledge transfer
    scheme (yellow) and by the EdgeBox+objectness baseline (red) on the target
    training set. Our technique steers localization towards complete objects
    and away from backgrounds. Labels are shown on the images.}
\end{figure*}

%% file: results_accross_datasets.tex
\section{Generalization across datasets}

The experiments in Sec.~\ref{cptResults} suggest a relatively easy recipe for knowledge transfer for WSOL:
train a strong class-generic proposal generator on a source training set with object bounding-boxes,
and use its proposals and scores inside MIL on a target set with only image-level labels. We
demonstrate here this recipe in several cross-dataset experiments, going beyond within-dataset
transfer typically shown in previous
works~\cite{guillaumin12cvpr,hoffman16jmlr,rochan2015cvpr,shi12bmvc,tang16cvpr}.

For these experiments, we switch to a stronger object proposal generator than SSD with Inception-V3: Faster-RCNN~\cite{ren15nips} with Inception-ResNet\cite{szegedy17aaai}.
Note that considering a single class-generic `entity' class avoids the technical problem raised in
Sec.~\ref{cptTransferLearning}, as Faster-RCNN will now output a single set of proposals, along with a single score (for objectness). The rest of our framework remains unaltered.

\begin{table}
  \footnotesize
  \hspace{-.2cm}
  \setlength\tabcolsep{1.5pt}
  \begin{tabular}{|l||c|c|c|c|c|c|}
    \hline
    \multirow{2}{*}{\diagbox{source set}{target set}}
    & \multicolumn{2}{c|}{ILSVRC target} & \multicolumn{2}{c|}{COCO 2014 train} &
    \multicolumn{2}{c|}{OID V2 val} \\
    \cline{2-7}
    & \hspace{.12cm} $> 0.5$ \hspace{.12cm} & $> 0.7$ & \hspace{.09cm} $> 0.5$ \hspace{.09cm} &  $> 0.7$ & $> 0.5$ &  $> 0.7$ \\
    \hline
    ILSVRC source & 74.2 & 61.7 & 34.5 & 26.8 & 62.0 & 51.8 \\
    COCO 2014 train & 67.7 & 58.5 & - & - & 59.5 & 49.8 \\
    PASCAL 2007 trainval & 59.5 & 47.2 & 26.2 & 20.8 & 55.3 & 42.2 \\
    \hline
    \hline
    EdgeBox + objectness & 50.5 & 28.5 & 20.6 & 10.2 & 32.4 & 16.3 \\
    \hline
  \end{tabular}
  \caption{MIL + Knowledge transfer across datasets: CorLoc results for IoU $> 0.5$ and $> 0.7$ on target
  datasets. Even knowledge transfer from the small PASCAL 2007 trainval works better than the
baseline of EdgeBoxes with objectness. Generally, transfer works better when the source training set
contains more classes.
Note that CorLoc when transferring from ILSVRC 2013 source train to ILSVRC 2013 target train is higher than in Tab.~\ref{tabResults} due to using a stronger proposal generator.}
  \label{tabAcrossDatasets}
  \vspace{-3mm}
\end{table}

As source training sets we use: (1) the ILSVRC 2013 source training set as defined in
Sec.~\ref{cptResults} (100 classes, 63k images), (2) the COCO 2014 training set~\cite{lin14eccv} (80
classes, 83k images), and (3) the PASCAL VOC 2007 trainval set~\cite{everingham10ijcv} (20 classes,
5011 images).
As target training sets we use (1) the ILSVRC 2013 target training set as defined in
Sec.~\ref{cptResults} (100 classes, 65k images), (2) the COCO 2014 training set~\cite{lin14eccv} (80
classes, 83k images), and (3) the Open Images V2 dataset~\cite{openimages_v2}, combining the
validation and test set~\cite{openimages_v2} (600 classes, 167k images).
In this experiment we do not try to remove source classes from the target training sets.

Tab.~\ref{tabAcrossDatasets} presents our across-dataset results and the MIL baseline using EdgeBoxes~\cite{dollar14eccv} with objectness (Sec.~\ref{cptMIL}).
We observe that the knowledge transfer method considerably outperform the baseline for all dataset pairs.  This is especially true at IoU $> 0.7$, where even using the small PASCAL VOC 2007 dataset as source yields 1.6-2.6$\times$ higher CorLoc than the baseline. Furthermore, using more source classes is consistently better for all target
datasets: ILSVRC 2013 (100 classes) is the best source, followed by COCO 2014 (80 classes), and then by PASCAL VOC 2007 (20 classes). This is despite COCO 2014 train having many more object instances (605k boxes) than the ILSVRC 2013
source train set (81k boxes).
We conclude that our knowledge transfer strategy works much better than the weakly supervised baseline and generalizes well across datasets.

%% file: conclusion.tex
\section{Conclusions}

We proposed a unified knowledge transfer framework for weakly supervised object localisation, which
enabled exploring knowledge transfer functions ranging from class-specific to class-generic.  Our
experiments on ILSVRC~\cite{russakovsky15ijcv} demonstrate:
(1) knowledge transfer at any level of generality substantially improve results, with class-generic knowledge transfer working best.
(2) class-generic knowledge transfer leads to large improvements over a weakly supervised baseline using manually engineering objectness~\cite{dollar14eccv}: +19.8\% CorLoc and +11.5\% mAP.
(3) our method delivers target class detectors reaching 80\% of the accuracy of their fully supervised counterparts.
(4) we outperform the best reported transfer learning
results on this dataset:
+41\% CorLoc and +3\% mAP over~\cite{hoffman16jmlr,tang16cvpr},
+16.2\% mAP over~\cite{redmon17cvpr}.
Moreover, across-dataset~\cite{lin14eccv,openimages_v2,russakovsky15ijcv} experiments demonstrate
(5) the general applicability of our technique.